\documentclass[sigconf]{acmart}

\AtBeginDocument{%
  \providecommand\BibTeX{{%
    \normalfont B\kern-0.5em{\scshape i\kern-0.25em b}\kern-0.8em\TeX}}}

\setcopyright{acmcopyright}
\copyrightyear{2022}
\acmYear{2022}
\acmDOI{10.1145/3510427.3510429}

\acmConference[ICBBB '22]{2022 12th International Conference on Bioscience, Biochemistry and Bioinformatics}{January 7--10, 2022}{Tokyo, Japan}
\acmBooktitle{2022 12th International Conference on Bioscience, Biochemistry and Bioinformatics (ICBBB '22), January 7--10, 2022, Tokyo, Japan}
\acmPrice{15.00}
\acmISBN{978-1-4503-8738-5/22/01}

\begin{document}

\title{Two Deep Learning Approaches for Automated Segmentation of Left Ventricle in Cine Cardiac MRI}

\author{Wenhui Chu}
\email{wchu@uh.edu}
\affiliation{%
  \institution{MRI Lab, University of Houston}
  \city{Houston}
  \state{TX}
  \country{USA}
}

\author{Nikolaos V. Tsekos}
\email{nvtsekos@central.uh.edu}
\affiliation{%
  \institution{MRI Lab, University of Houston}
  \city{Houston}
  \state{TX}
  \country{USA}
}

\begin{abstract}
Left ventricle (LV) segmentation is critical for clinical quantification and diagnosis of cardiac images. In this work, we propose two novel deep learning architectures called LNU-Net and IBU-Net for left ventricle segmentation from short-axis cine MRI images. LNU-Net is derived from layer normalization (LN) U-Net architecture, while IBU-Net is derived from the instance-batch normalized (IB) U-Net for medical image segmentation. The architectures of LNU-Net and IBU-Net have a down-sampling path for feature extraction and an up-sampling path for precise localization. We use the original U-Net as the basic segmentation approach and compared it with our proposed architectures. Both LNU-Net and IBU-Net have left ventricle segmentation methods: LNU-Net applies layer normalization in each convolutional block, while IBU-Net incorporates instance and batch normalization together in the first convolutional block and passes its result to the next layer. Our method incorporates affine transformations and elastic deformations for image data processing. Our dataset that contains 805 MRI images regarding the left ventricle from 45 patients is used for evaluation. We experimentally evaluate the results of the proposed approaches outperforming the dice coefficient and the average perpendicular distance than other state-of-the-art approaches.
\end{abstract}




\maketitle

\section{Introduction}

Cardiac diseases are a consistent threat to the lives of people everywhere. In the US, this is the most significant cause of natural death, which leads to the loss of millions of lives every year \cite{cdc}. In U-Net, it provides precise segmentation for margin detection, and this is especially important for clinical application. In addition, U-Net is efficient in using GPU memory. Because GPU memory is a bottleneck compared to its computation power, saving GPU memory is quite a significant advantage for CNN. U-Net also outperformed related work because it required fewer annotated training samples. As we know, it is quite expensive to acquire annotated samples, so U-Net is excellent in getting high accuracy with less training data.

MRI images of cardiac diseases are an important reference for the diagnosis of heart diseases. Manual segmentation of MRI images requires plenty of time and the additional expenses of employing experts, which could be a valuable resource for more patients. In addition, this form of segmentation's accuracy cannot be fully guaranteed. With the development of computational units, neural network scans can be applied in almost every aspect of our daily life. Medical care is also an important application, especially for image classification, image segmentation, and objective detection. It is believed that image processing is more efficient and accurate with the help of neural networks. Specifically, for the left ventricle cardiac image, we think a more accurate segmentation will be helpful for the diagnosis of heart disease, and this will save more lives in the future. Because accurate segmentation is the very first step in the evaluation of cardiac function, there have been some important studies on the image segmentation for left ventricle cardiac in the past several decades \cite{petitjean2011review}. In this paper, we focus on pixel classification methods with U-Net \cite{ronneberger2015unet}, which is a neural network widely used for medical applications. The U-Net will be improved with different normalization methods, and we will compare their effects on results.

In the past several years, efforts have been made for the application of neural networks in image segmentation. R. Poudel et al. \cite{poudel2016recurrent} used recurrent fully convolutional neural networks for MRI cardiac image segmentation. A fully convolutional neural network was applied for cardiac segmentation \cite{tran2016fully}. Left ventricle segmentation from cardiac MRI was processed with the combination of level set methods and deep belief networks \cite{ngo2013left}. With a deep learning network, sufficient training data is a necessity for improved accuracy \cite{queiros2014fast}. Before that, machine learning methods, such as the Gaussian-mixture model \cite{hu2013hybrid}, image-based approaching \cite{huang2011image}, layer spatiotemporal forests \cite{margeta2012layered}, and dynamic methods \cite{liu2012automatic} were applied for this task.

In this study, we would like to compare the effects of different normalization methods with U-Net in the application of left ventricle cardiac image segmentation. The paper will be organized as follows. In the section on methodology, we will discuss the reason for using U-Net, and make a brief comparison of the normalization methods we use, including batch normalization, layer normalization, and batch instance normalization. We will display the structure of the neural networks and the data we use. In section III, we will discuss the methods of data preprocessing and the experiment results. The last section of the paper will be the discussion and conclusion.

\section{Methodology}

\subsection{The Structure of Networks}

U-Net is a convolutional neural network (CNN) specially designed for biomedical image segmentation \cite{ronneberger2015unet}. U-Net has a U-shaped architecture and we will discuss it in this section. There were some variants based on U-Net, including UNet++ \cite{zhou2018unetpp}, Attention UNet \cite{oktay2018attention} and BNU-Net \cite{chu2019bnu}.

We use three normalization methods in this study, which are batch normalization \cite{ioffe2015batch}, layer normalization \cite{ba2016layer}, and instance normalization \cite{ulyanov2016instance}. They can generally be viewed as a technique for data preprocessing. Normalization is applied for deep learning in order for more accurate and efficient training. For the improvement of accuracy, it is especially effective when features are widely distributed because all of the features are forced into a similar range. It also helps to reduce the internal covariate shift, which means it weakens the changes in network parameters during training. The larger the difference in feature range, the more effectively the data normalization method will restrict the weights into a certain range and prevent them from exploding. Therefore, the optimization process will become faster.

\begin{figure}[t]
  \centering
  \includegraphics[width=\linewidth]{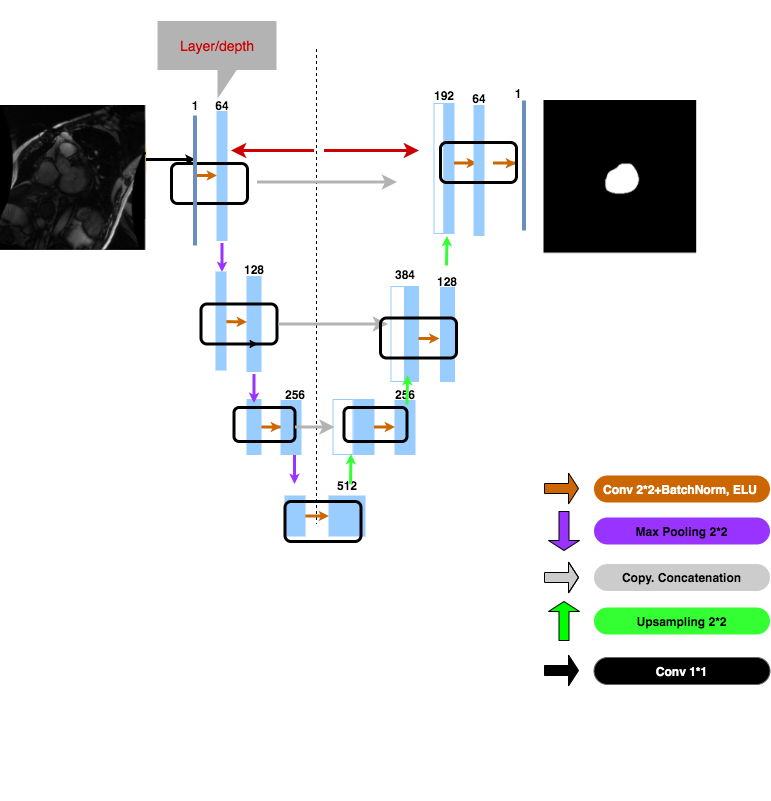}
  \caption{Architecture of the BNU-Net convolutional network. (a) The contraction path is responsible for feature extraction. (b) Batch normalization is performed after each convolution in the convolutional layer.}
  \label{fig:bnu}
\end{figure}

Batch normalization \cite{ioffe2015batch} is the first method we test. It normalizes the output of previous activations across the batch in a network. You can view it as doing data pre-processing at each layer of the neural network. When we use batch normalization, we insert a batch norm layer into the U-Net. See Figure \ref{fig:bnu}.

Batch normalization has a reputation for speeding up the training as well as the generalization of convolution neural networks(CNN) \cite{ioffe2015batch}. However, for recurrent architectures, this technique fails greatly since its application is only limited to stacks of RNN, where the application of normalization is vertical to avoid the problem of repeated rescaling.

Batch normalization harbors a few weaknesses that limit its ability to be considered the most influential technique to reduce the internal covariant shift (ICS) in deep neural network models \cite{ioffe2015batch}. During the normalization of outputs from previous models, the batch is divided by the empirical standard deviation, and the result is subtracted from the subsequent empirical mean. Batch normalization is typically poor in the pipelining of online learning, thereby contributing to poor generalization of the training data that is contributed by the change in batch size which each iteration. This resulting shift in the input data eventually affects its performance. To solve this problem, other normalization methods were proposed.

\begin{figure}[t]
  \centering
  \includegraphics[width=\linewidth]{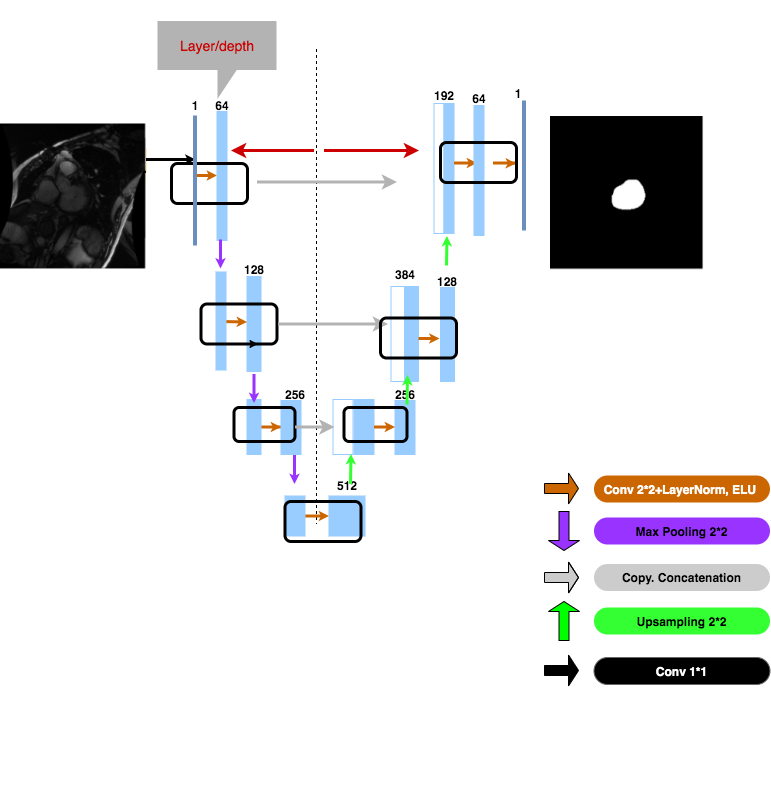}
  \caption{The LNU-Net architecture of the proposed fully convolutional network. Layer normalization is performed after each convolution in the convolutional layer.}
  \label{fig:lnu}
\end{figure}

Layer normalization \cite{ba2016layer} normalizes the input across features, which is completely different from normalizing across batches in batch normalization. The calculation for mean and standard deviation is similar to that in batch normalization. The mean across the features and variance across the features are calculated for each element of the input. We applied layer normalization in each convolutional block and based it on an exponential linear unit (ELU). See Figure \ref{fig:lnu}.

In batch normalization, the statistics are the same for each batch, while in layer normalization, the statistics are computed across each feature and are independent of other examples. The independence between examples makes it simple when applied to recurrent neural networks. The authors of ``Recurrent Fully Convolutional Neural Networks for Multi-slice MRI Cardiac Segmentation'' \cite{poudel2016recurrent} found it outperformed other normalization methods in the application or RNNs.

\begin{figure}[t]
  \centering
  \includegraphics[width=\linewidth]{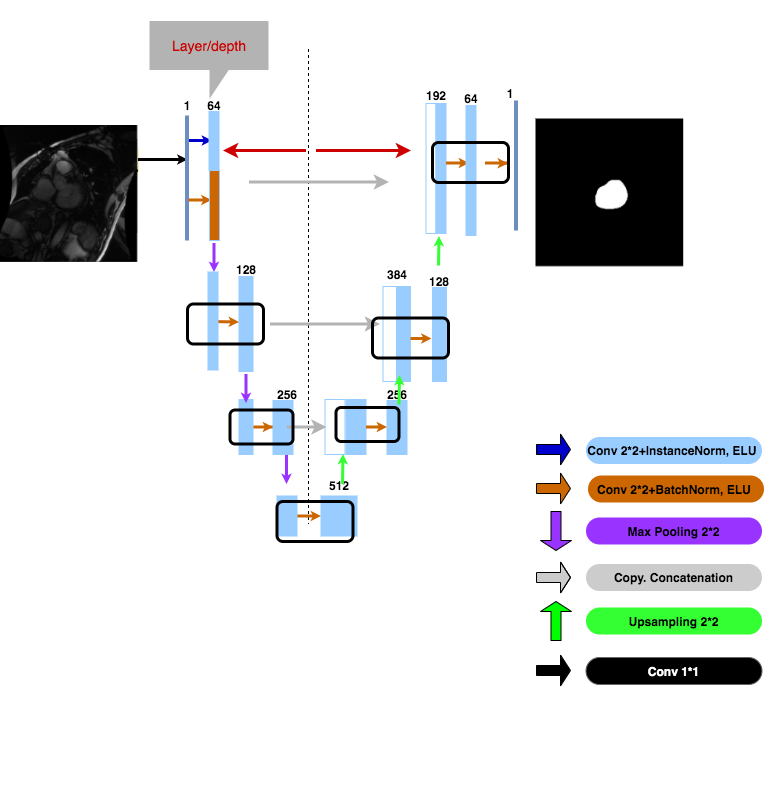}
  \caption{Architecture of the IBU-Net convolutional network. Instance normalization is applied in the first convolutional layer. Batch normalization is performed after each convolution in the convolutional layer.}
  \label{fig:ibu}
\end{figure}

Batch instance normalization \cite{nam2018batch} is an interpolation of batch normalization and instance normalization \cite{ulyanov2016instance}. With instance normalization, the mean and standard deviation are computed across each channel for each input group. Instance normalization shows some similarity with layer normalization, but there is one obvious difference: layer normalization calculates the statistics across each feature in the training sample, while instance normalization calculates across each channel. In recent years, instance normalization was also found to be effective for Generative Adversarial Networks (GANs). However, there are also problems with instance normalization: it is suitable for style transfer cases, but incapable of contrast matters. This problem makes it unsuitable for image segmentation tasks. To solve this problem, we use instance and batch normalization together in the first convolutional block, apply an ELU and pass its result to the next layer. See Figure \ref{fig:ibu}. It balances the batch normalization and instance normalization, and the model could learn to use different combination percentages with gradient descent. According to the previous work \cite{nam2018batch}, batch instance normalization outperforms batch normalization on CIFAR-10/100, ImageNet, adaptation, style transfer, and image classification tasks.

\subsection{Dataset}

The Sunnybrook dataset \cite{sunnybrook} was used for the experiments. It was received from Imaging Research Centre for Cardiovascular Intervention in Sunnybrook Health Sciences Centre and specially made for automatic left ventricle segmentation for MICCAI 2009. The ground truth masks were provided by experts that were manually segmented. Medical images from 45 patients (a total of 805 images) were collected in the Sunnybrook dataset, including different types of cardiac conditions: heart failure with infarction (12 cases), heart failure without infarction (12 cases), hypertrophy (12 cases), and healthy patients (9 cases). Each patient has 12 to 28 images. Each time series consists of 6 to 12 2D cine stacks with 8 mm slice thickness and 1.3mm to 1.4 mm in-plane resolution. This dataset split the training, validation, testing sets in the ratio of 15 : 15 : 15. The experiment was conducted with an NVIDIA GeForce Titan X Pascal GPU. When we compare the computational speed for each epoch, the performance of computational resources should also be considered.

\subsection{Model}

In this study, the model we used for image segmentation was a modified U-Net based on a fully convolutional network. We added Exponential Linear Units (ELU) as an activation function instead of ReLU in the original version. In addition, we also added a normalization layer into the model, which was not considered in the original U-Net. You can refer to Figures \ref{fig:bnu}-\ref{fig:ibu}, to the architecture of models.

\begin{itemize}
\item Conv2*2 is a convolution layer with a 2*2 kernel.
\item ELU is applied instead of ReLU to speed up the training process. It also avoids the vanishing gradient problem, because the value is no longer 0 with a negative x.
\item Normalization: Normalized feature map to solve internal covariate shift.
\item Cropping2D is a cropping layer designed for cropping feature maps. It is also used to reduce concatenating to avoid overfitting.
\item Concatenating is two connecting different feature maps from downsampling.
\item UpSampling layer is used to increase the size of the feature.
\end{itemize}

As we see, the new U-Net for automated left ventricle segmentation is based on the basic structure of a fully convolutional neural (FCN) network architecture \cite{shelhamer2017fully}. The first improvement we made was an encoder architecture. It is a computationally expensive process to train an FCN for medical image segmentation, because of the large number of multiply-and-adds required for image processing. An encoder architecture would help to ease the computational burden and quicken the training process, especially when the computational power of Titan X was limited.

The second issue is the improvement of normalization. As we mentioned, three normalization methods are attempted for model improvement, which are batch normalization, layer normalization, and batch instance normalization. We found that for medical image segmentation, internal covariate shift is a problem during training with an FCN. This was encumbered by the distribution of input feature change \cite{zhou2019normalization}. It also leads to a slow down in training. Because training was a process independent of initial data distribution, we thought it was a good attempt to involve data normalization in our study.

The third issue is to skip connection in FCN. We found it lacked well-defined edges with FCN segmentation. The boundaries between the images were not clearly defined with the original version of FCN.

\section{Results}

\subsection{Experimental Studies}

In order to achieve great learning performance, we need a large number of the labeled data set. However, in current medical areas, there are limited training samples. Therefore, the data augmentation methods are applied in medical image datasets. Patrice et al \cite{simard2003best} first proposed the elastic deformation in 2003. We use elastic deformation to the available training images that increase the size of the training dataset and improve the adaptability of the models. In this work, we use the following data augmentation strategies: affine transformation, elastic deformation, and rotation.

The original version of U-Net is a typical method for left ventricle cardiac image, so we first test its performance with an NVIDIA GeForceTitan X Pascal GPU. We found that the dice mean was only 0.87 with U-Net. It took 11s per epoch on training if we set the batch size as 16. In this study, a novel segmentation schema that contains batch normalization and instance normalization was proposed to improve the segmentation performance. We want to improve the dice mean and to accelerate the training process.

\subsection{Experimental Results}

\begin{table}[t]
\caption{Dice Mean with Different Normalization}
\label{tab:dice_norm}
\begin{tabular}{lcc}
\toprule
 & With ELU & With ReLU \\
\midrule
Instance-Batch Normalization & 0.94 & 0.93 \\
Batch Normalization & 0.91 & 0.90 \\
Layer Normalization & 0.89 & 0.88 \\
\bottomrule
\end{tabular}
\end{table}

We use U-Net with batch normalization, U-Net with layer normalization, and U-Net with batch-instance normalization to deal with image segmentation tasks. The dice mean are listed in Table \ref{tab:dice_norm}. ELU was likely to introduce more calculations compared to ReLU. But when we had a negative x, ELU would avoid the gradient vanishing led by ReLU. Table \ref{tab:dice_norm} displays the dice mean for six experiments: batch normalization with ELU and ReLU, layer normalization with ELU and ReLU, as well as batch instance normalization with ELU and ReLU. We would like to test the influence of different activation functions on image segmentation, on both dice mean and training speed. In the meanwhile, we also wanted to know whether normalization methods would lead to better performance, and which normalization was optimal in this case.

From the experiment results, we can see that the new activation function, ELU, was effective in increasing the training dice mean. In each comparison, the model performance with ELU was better than that with ReLU. It was agreed that ELU was more suitable for cardiac MRI image segmentation.

\begin{table*}[t]
\caption{Output of the Models and Efficiency Metrics Results}
\label{tab:models}
\begin{tabular}{lcccc}
\toprule
 & Dice mean & Dice std & Sensitivity & Average perpendicular distance \\
\midrule
IBU-Net with data augmentation & \textbf{0.96} & \textbf{0.02} & \textbf{0.98} & \textbf{1.91} \\
IBU-Net without data augmentation & 0.94 & 0.03 & 0.96 & 2.02 \\
LNU-Net with data augmentation & 0.90 & 0.11 & 0.96 & 2.29 \\
LNU-Net without data augmentation & 0.89 & 0.14 & 0.95 & 2.46 \\
BNU-Net \cite{chu2019bnu} with data augmentation & 0.93 & 0.03 & 0.97 & 1.94 \\
BNU-Net \cite{chu2019bnu} without data augmentation & 0.91 & 0.04 & 0.96 & 2.06 \\
U-Net with data augmentation & 0.88 & 0.09 & 0.96 & 2.48 \\
U-Net without data augmentation & 0.87 & 0.11 & 0.95 & 2.51 \\
\bottomrule
\end{tabular}
\end{table*}

The performance of LNU-Net (Figure \ref{fig:lnu}) and IBU-Net (Figure \ref{fig:ibu}) is shown in Table \ref{tab:models}. This table illustrates of LNU-Net and IBU-Net varies with and without data augmentation. When comparing three normalization methods in Table \ref{tab:models}, we found that batch instance normalization is the best choice for image segmentation. As we mentioned in Section II, batch instance normalization combined the benefits of batch normalization and instance normalization, getting better performance in image segmentation tasks.

Compared to the original version of U-Net, the combination of ELU and batch instance normalization improved the dice mean by 8\% from 0.87 to 0.96, which is a significant and reliable improvement. The improvement in training speed was not displayed in the table above. Each epoch was trained with 11 seconds in the original U-Net. While with the combination of encoder architecture, batch instance normalization, and drop-connection, it only took 8 seconds to train an epoch. The training speed was improved by 27\%, which meant almost one-third more images could be processed within the same period. The acceleration was due to the encoder architecture and drop-connection. These could reduce the calculation during training, and they could also avoid overfitting.

\begin{figure*}[t]
  \centering
  \includegraphics[width=\textwidth]{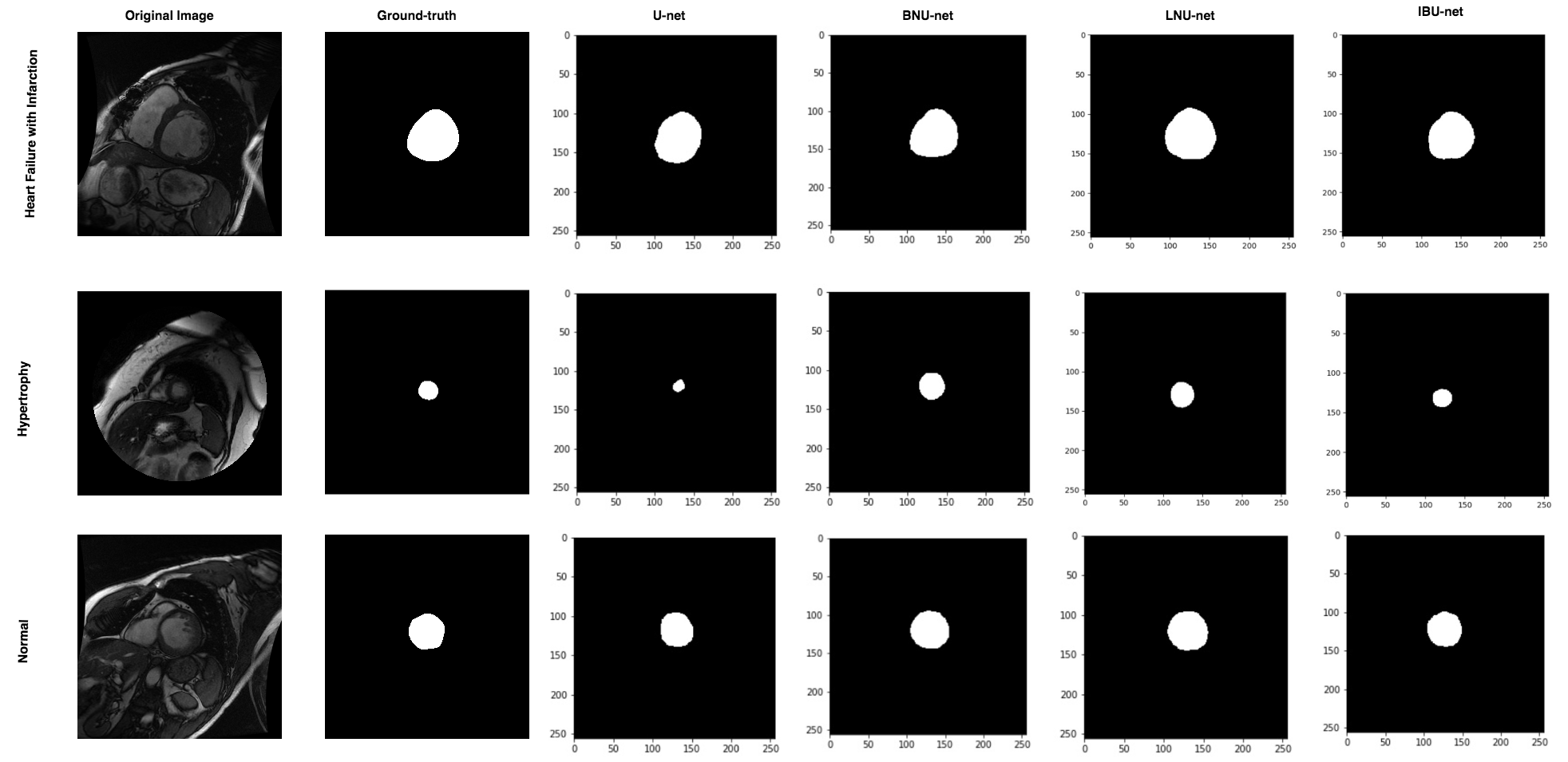}
  \caption{Examples of segmentation results on raw inputs from three conditions in Sunnybrook dataset. The first row contains heart failure with infarction, the second row represents hypertrophy, the third row shows healthy patients.}
  \label{fig:seg_results}
\end{figure*}

\begin{figure*}[t]
  \centering
  \includegraphics[width=\textwidth]{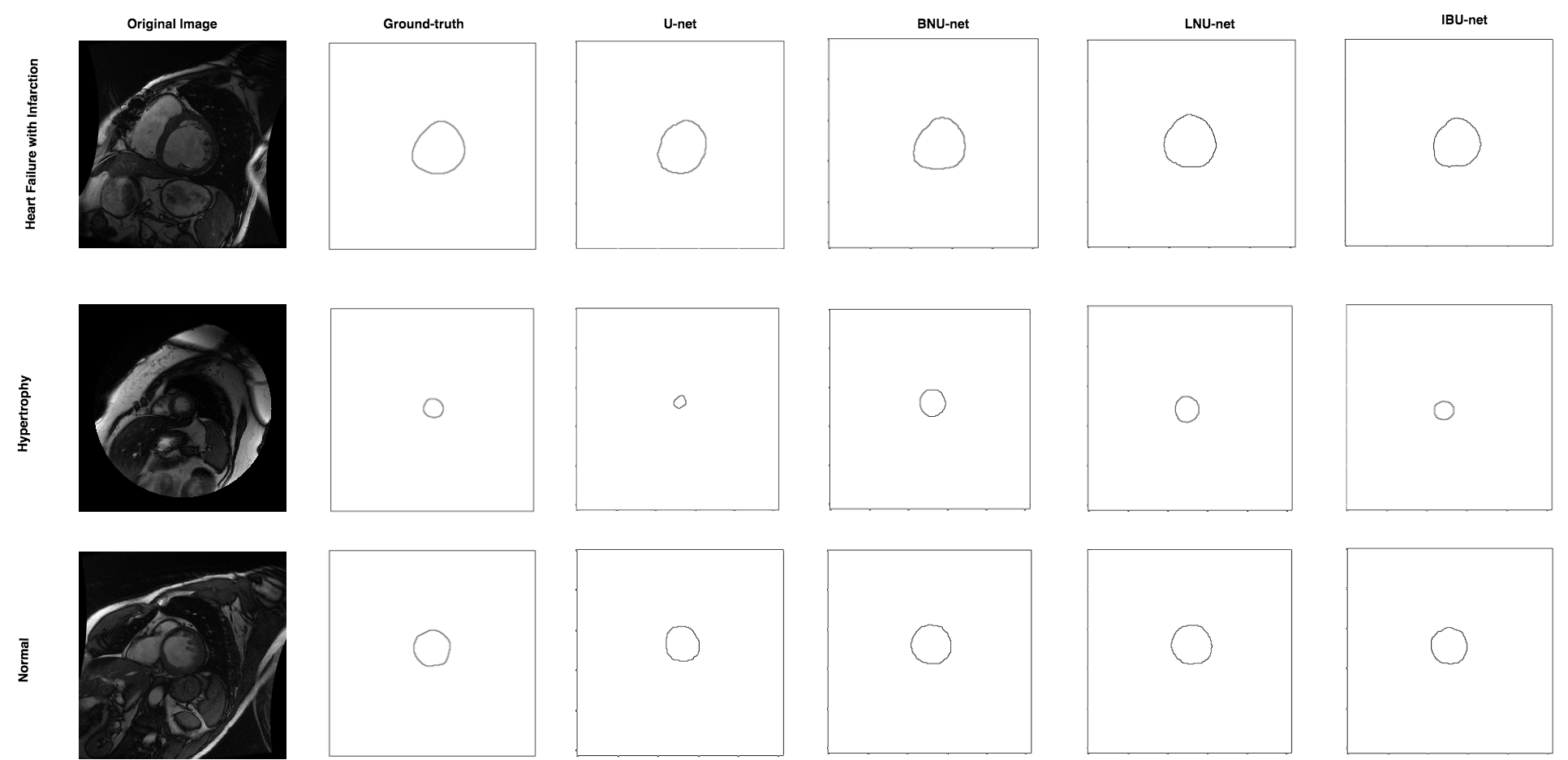}
  \caption{Some segmentation outputs by our methods. The solid lines represent the segmentation examples of Sunnybrook dataset. We compare four different network methods, which are U-Net, BNU-Net, LNU-Net, and IBU-Net respectively.}
  \label{fig:seg_outputs}
\end{figure*}

Figure \ref{fig:seg_results} and Figure \ref{fig:seg_outputs} show three examples of the output segmentation suffering from different cardiac conditions: heart failure with infarction, hypertrophy, and healthy patients. To evaluate the performance of our proposed structures, we use the same Sunnybrook dataset and same data split, compare the performance of our methods with other groups report their performances on the 45 patient cases. The amount of performance improvement varies across different normalization methods. In Table \ref{tab:comparison}, the observation is that using IBU-Net with data augmentation achieves the best dice mean, dice std, and average perpendicular distance when comparing with other groups' proposed structures.

\begin{table}[t]
\caption{Dice Score and Average Perpendicular Distance (APD) of Segmenting the Sunnybrook Dataset, Compared to the Performance from the State of the Art Methods}
\label{tab:comparison}
\begin{tabular}{lccc}
\toprule
 & Dice Mean & Dice Std & APD (mm) \\
\midrule
IBU-Net & \textbf{0.96} & \textbf{0.02} & \textbf{1.91} \\
X-Y Zhou \cite{zhou2019normalization} & 0.92 & - & - \\
Zhou \cite{zhou2019unet} & 0.93 & 0.06 & - \\
Ngo and Carneiro \cite{ngo2013left} & 0.90 & 0.03 & 2.08 \\
Hu et al \cite{hu2013hybrid} & 0.89 & 0.03 & 2.24 \\
Huang et al \cite{huang2011image} & 0.89 & 0.04 & 2.16 \\
Liu et al \cite{liu2012automatic} & 0.88 & 0.03 & 2.36 \\
\bottomrule
\end{tabular}
\end{table}

The experimental results for the algorithm above indicate that our proposed methodology has improved efficiency and effectiveness, and outperforms convolutional deep networks contributed by the increase in the resultant dice score.

\section{Discussion and Conclusion}

In this study, we improved the left ventricle cardiac image segmentation tasks with a renewal U-Net. Data preprocessing was conducted with batch normalization, layer normalization, and batch-instance normalization. Encoder architecture and drop-connection were also conducted to accelerate training and reduce over-fitting. The major contributions by modifying U-Net in this study are listed as follows.

Firstly, the normalization method was effective for improving the dice score in image segmentation tasks. Among the widely-used normalization methods, we found that batch instance normalization was better to perform segmentation. With the combination of batch instance normalization and ELU, we got a state-of-the-art segmentation dice mean of 0.96. Secondly, the training speed was accelerated with our current methods. The training time per epoch was reduced from 12s to 9s. Thirdly, compared to the original version of U-Net, our version displayed a higher dice score and faster speed, which means we got better results with less computational resources. Cardiac CINE magnetic resonance imaging is the gold standard for the assessment of cardiac function \cite{kustner2020cinenet}. We hope our work could be helpful for the practical image segmentation work in medical applications. We intend for this work to relieve both experienced doctors and recovering patients in their fight against cardiac diseases.

\bibliographystyle{ACM-Reference-Format}
\bibliography{references}

\end{document}